\pdfoutput=1
\documentclass[11pt]{article}

\usepackage[]{acl}
\usepackage{multirow}
\usepackage{times}
\usepackage{latexsym}
\usepackage{graphicx}
\usepackage{enumitem}
\usepackage[T1]{fontenc}
\usepackage[utf8]{inputenc}

\usepackage{microtype}
\usepackage{siunitx}[=v2]
\usepackage{mathtools}

\title{ParrotTTS: Text-to-speech synthesis exploiting \\ disentangled self-supervised representations}

\author{%
 $^{*}$Neil Shah$^{1,2}$ \mbox{ }
 $^{*}$Saiteja Kosgi $^{1}$  \mbox{ } \textbf{Vishal Tambrahalli} $^{1}$ \mbox{ }  \textbf{Neha Sahipjohn}$^{1}$ \mbox{ } \\
 \textbf{Niranjan Pedanekar}$^{2}$ \mbox{ }\textbf{Vineet Gandhi}$^{1}$ \vspace{4mm} \\ 
 $^{1}$Kohli Centre on Intelligent Systems, IIIT Hyderabad \mbox{ } \\ $^{2}$TCS Research, Pune \mbox{ } \\
 \small{\texttt{\{saiteja.k,neil.shah,vishal.tambrahalli,neha.s\}@research.iiit.ac.in}} \mbox{  }\\
\small{\texttt{vgandhi@iiit.ac.in}} \\
}

\newcommand{\ourmodel}{ParrotTTS}
\newcommand{\spencoder}{STE}
\newcommand{\spdecoder}{ETS}
\newcommand{\txencoder}{TTE}
\newcommand{\khz}{kHz}
\newcommand{\hmb}[1]{\mathbf{h}_{#1}}
\newcommand{\ar}{AR}
\newcommand{\nar}{NAR}
\newcommand{\baselineml}{MetaTTS}

\begin{document}
\maketitle
\begin{abstract}
\footnote{* equal contribution}
% ACL: 
% - new modular architecture [refer]
% - improved data efficiency [claim]
% - better multispeaker adaptation [claim]
% - stabler training, faster inference, smaller mem footprint [claim]
% INT: 
% - new and novel language adaptation including low-resource [claim]
% - better inter-language voice cloning [claim]
%We present \ourmodel, a modularized text-to-speech (TTS) synthesis model exploiting disentangled self-supervised speech representations. We show that transcripts from a single speaker are sufficient to train its multi-speaker variant. 

We present \ourmodel, a modularized text-to-speech synthesis model leveraging disentangled self-supervised speech representations. It can train a multi-speaker variant effectively using transcripts from a single speaker. \ourmodel~adapts to a new language in low resource setup and generalizes to languages not seen while training the self-supervised backbone. Moreover, without training on bilingual or parallel examples, \ourmodel~can transfer voices across languages while preserving the speaker-specific characteristics, e.g., synthesizing fluent Hindi speech using a French speaker’s voice and accent. We present extensive results in monolingual and multi-lingual scenarios. \ourmodel~outperforms state-of-the-art multi-lingual TTS models using only a fraction of paired data as latter. Speech samples from \ourmodel~can be found at \textcolor{blue}{\url{https://parrot-tts.github.io/tts/}}

\end{abstract}

% Text-to-speech (TTS) systems were modelled as mel-synthesizers followed by speech-vocoders since the era of statistical TTS that is carried forward into neural designs. We propose an alternative approach to TTS modelling referred to as \ourmodel~borrowing from self-supervised learning (SSL) methods. \ourmodel~takes a two-step approach by initially training a speech-to-speech model on unlabelled data that is abundantly available, followed by a text-to-embedding model that leverages speech with aligned transcriptions to extend it to TTS. \ourmodel~achieves competitive mean opinion scores on naturalness compared to traditional TTS models but significantly improves over the latter's data efficiency of transcribed pairs and speaker adaptation without transcriptions.
% This further paves the path to training TTS models on generically trained SSL speech models. Speech samples from \ourmodel~can be found at \textcolor{blue}{\url{https://parrottts.github.io/tts/}} 
% \end{abstract}

\section{Introduction}

Vocal learning forms the first phase of infants starting to talk~\cite{locke1996infants,locke1994phases} by simply listening to sounds/speech.
It is hypothesized~\cite{kuhl1996infant} that infants listening to ambient language store perceptually derived representations of the speech sounds they hear, which in turn serve as targets for the production of speech utterances. Interestingly, in this phase, the infant has no conception of text or linguistic rules, and speech is considered sufficient to influence speech production~\cite{kuhl1996infant} as can parrots~\cite{locke1994phases}.

% Vocal learning forms the first phase of infants starting to talk~\cite{locke1996infants,locke1994phases}. In this phase, learning happens by simply listening to sounds/speech. Studies show that vocal learning begins in the final trimester of pregnancy; the normally developing fetus can hear its mother's voice within the womb~\cite{kolata1984studying}. Several studies show that the best way to promote language development for babies is to talk to them. It is hypothesized~\cite{kuhl1996infant} that infants listening to ambient language store perceptually derived representations of the speech sounds they hear, which in turn serve as targets for the production of speech utterances. Interestingly, in this phase, the infant has no conception of text or linguistic rules, and speech is considered sufficient to influence speech production~\cite{kuhl1996infant}. Eventually, if parrots can talk without understanding language, there is no reason human infants should need to possess grammatical capability either to comprehend and produce speech~\cite{locke1994phases}.

\begin{figure}[t]
  \centering
\includegraphics[width=0.9\linewidth]{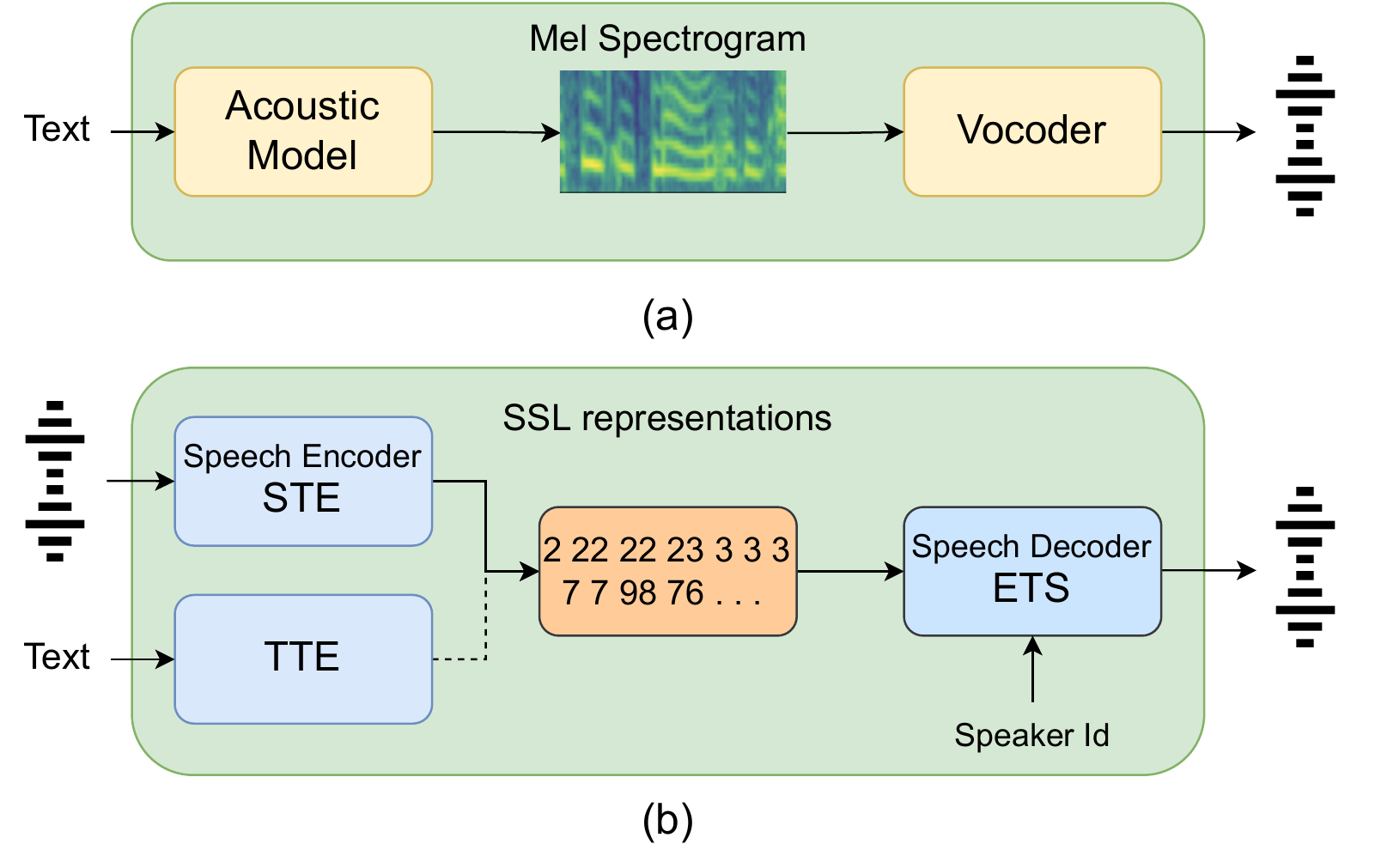}
\caption{ (a) Traditional mel-based TTS and (b) Proposed TTS model} 
  \label{fig:teaser}
\end{figure}

Our proposed \ourmodel~model follows a similar learning process. 
Our idea mimics the two-step approach, with the first learning to produce sounds capturing the whole gamut of phonetic variations.
It is attained by learning quantized representations of sound units in a self-supervised manner using the raw audio data. The second phase builds on top of the first by learning a content mapping from text to quantized speech representations (or embeddings). Only the latter step uses paired text-speech data. The two phases are analogous to first \textit{learning to talk} followed by \textit{learning to read}. 

Figure~\ref{fig:teaser}~illustrates \ourmodel~contrasting it with the traditional mel-based TTS. The SSL module includes a speech-to-embedding (\spencoder) encoder trained on masked prediction task to learn an embedding representation of the input raw audio~\cite{baevski2020wav2vec,hsu2021hubert,van2017neural}.
An embedding-to-speech (\spdecoder) decoder is independently trained to invert embeddings to synthesize audio waveforms and is additionally conditioned on speaker identity. This \textit{learning to talk} is the first of the two-step training pipeline. In the subsequent \textit{learning to read} step, a separate text-to-embedding (\txencoder) encoder is trained to generate embeddings from text (or equivalent phonetic) inputs. This step requires labeled speech with aligned transcriptions.

\ourmodel~offer several advantages over the traditional mel-based neural TTS models~\cite{ren2020fastspeech,wang2017tacotron}. For instance,  (a) Quantized speech embedding has lower variance than that of Mel frames reducing the complexity to train \txencoder~
%~and increasing capacity of downstream \spdecoder. 
(b) Direct waveform prediction bypasses potential vocoder generalization issues~\cite{kim2021conditional}. (c) Reduced complexity helps in stabler training of \txencoder~encoder for either autoregressive or non-autoregressive choice. For example, we observe at least eight-fold faster convergence in training iterations of our \txencoder~module compared to that of~\citet{ren2020fastspeech} and~\citet{wang2017tacotron}.

%We observe that, for example, our embeddings are speaker agnostic, requiring \spdecoder~conditioning on speaker identity for speaker adaptation.
%similar to FastSpeech2s~\cite{ren2020fastspeech}

While our work closely relates with recent works~\cite{du2022vqtts,wang2023neural,siuzdak2022wavthruvec} utilizing self-supervised representations for text-to-speech synthesis, our focus differs by aiming to achieve a unified multi-speaker, multi-lingual TTS system in low-resource scenarios \cite{xu2020lrspeech}. In our work, low-resource refers to the scarcity of paired text-to-speech data. Here are the key distinctions of our model compared to recent efforts:

%Our model distinguishes itself from recent efforts in the following ways:
% Our work is closely relates to recent efforts~\cite{du2022vqtts,wang2023neural,siuzdak2022wavthruvec} that employ self-supervised audio representations for the purpose of text-to-speech synthesis. Nevertheless, our focus diverges from these efforts as we strive to develop a unified multi-speaker, multi-lingual TTS in low-resource scenarios. The term "low-resource" in our work refers specifically to the limited availability of paired text-to-speech data. Our model stands out from previous endeavors in the following aspects:

% Our work closely relates with recent efforts~\cite{du2022vqtts,wang2023neural,siuzdak2022wavthruvec} utilizing self-supervised audio representations for text-to-speech synthesis. However, our focus is different and aims to learn an unified multi-speaker, multi-lingual TTS in low resource scenario. The low-resource in our work is in context of the availability of paired text-to-speech data. \ourmodel~contrasts with prior efforts in following ways:

%%ANIL do we merge point 1 with point 3 and specially call out few-shot (5h) language learning?

\begin{itemize}[leftmargin=*]
    \item Unlike contemporary efforts concentrated on large scale training~\cite{wang2023neural}, we focus on low resource adaptation.  \vspace{-0.7em} 
    \item We employ disentangled self-supervised representations~\cite{polyak2021speech} paired with independently trained \spencoder. This allows us to train multi-speaker TTS using paired data from a single speaker and still adapt it to novel voices with untranscribed speech alone. In contrast, prior efforts either limit to a single speaker TTS~\cite{du2022vqtts} or require paired text-audio data from multiple speakers during training~\cite{siuzdak2022wavthruvec}. \vspace{-0.7em}
    \item We show that the \ourmodel~can be extended to a new language with as little as five hours of paired data from a single speaker. The model generalizes to languages unseen during the learning of self-supervised representation. \vspace{-0.7em}
    \item Moreover, without training on any bilingual or parallel examples, \ourmodel~can transfer voices across languages while preserving the speaker-specific characteristics. We present extensive results on six languages in terms of speech naturalness and speaker similarity in parallel and cross-lingual synthesis. 
\end{itemize}

% We would like to emphasize that we do not claim general superiority over the other SSL based TTS models~\cite{}. The main contribution of this work is favorable results in the studied low resource scenario, where only small amount of paired data from a single speaker per language is available for training the TTS. We would also like point out that several methods~\cite{wang2023neural} partially/fully rely on Automatic Speech Recognition (ASR) for obtaining the paired data. It must be noted that these ASR models are trained using large amount of supervised data themselves, which may not be available in the low resource settings. 

Additionally, it's worth mentioning that certain methods~\cite{wang2023neural} depend partially or entirely on Automatic Speech Recognition (ASR) to obtain paired data. It should be noted that these ASR models are trained using substantial amounts of supervised data, inaccessible in low resource settings. 
% Finally, we would like to emphasize that we do not assert overall superiority compared to recent SSL-based TTS \cite{wang2023neural,siuzdak2022wavthruvec}. 
While architecturally similar to other SSL-based TTS~\cite{wang2023neural,siuzdak2022wavthruvec}, our primary contribution lies in achieving promising outcomes in the low resource scenario, where minimal paired data from a single speaker per language is accessible for TTS training. 
% The rest of the paper is organized as follows: our approach of modelling TTS using SSL is described in Section~\ref{sec:model}. We train multiple models of our \ourmodel~approach with different choices and study their effects like the quality of rendered speech, word-error rates upon transcription of speech output, etc., see Section~\ref{sec:experiments}. Experimental results reported in Section~\ref{sec:results} consistently point to the competitive or superior performance of \ourmodel~relative to the current state-of-the-art for TTS in both single and multi-lingual scenarios. While these observations are of significant value to practitioners in evaluating the adoption of \ourmodel~approach for speech synthesis, numerous questions need further investigation. We conclude in Section~\ref{sec:conclusion} with a discussion of these questions and the related topics that need further exploration to better understand the proposed approach.

% Overall our work makes following contributions: 
% \begin{itemize}
%     \item We propose ParrotTTS a novel formulation for text to speech synthesis. The primary novelty is in the training procedure, where a TTH module is learnt over pre-trained and frozen self-supervised embeddings. 
%     \item We perform thorough experiments to demonstrate the efficacy of ParrotTTS for new speakers, attaining quality of fully supervised models without using any transcripts. We present thorough qualitative results for text-to-speech synthesis and speech resynthesis.
% \end{itemize}

\section{Related work}
% TTS systems have been studied for decades now, with the concatenative statistical models from earlier attempts~\citep{hunt1996unit,cohn2020perception} being increasingly replaced by neural variants in recent years~\cite{oord2016wavenet}.
%We review the foundational neural TTS models in Section~\ref{sec:neuraltts}, unsupervised methods in Section~\ref{sec:unsuptts} their self-supervised counterparts in Section~\ref{sec:selftts} and multi-lingual TTS in section~\ref{sec:multitts}.

\subsection{Foundational Neural TTS models}
\label{sec:neuraltts}

Traditional neural TTS model encodes text or phonetic inputs to hidden states, followed by a decoder that generates Mels from the hidden states. 
Predicted Mel frames contain all the necessary information to reconstruct speech~\citep{griffin1984signal} and an independently trained vocoder~\citep{oord2016wavenet,kong2020hifi} transforms them into time-domain waves. Mel predicting decoders could be autoregressive/sequential ~\citep{wang2017tacotron,valle2020flowtron,shen2018natural} or non-autoregressive/parallel~\citep{ren2019fastspeech,ren2020fastspeech,lancucki2021fastpitch}. Non-autoregressive models additionally predict intermediate features like duration, pitch, and energy for each phoneme. They are faster at inference and robust to word skipping or repetition errors~\cite{ren2020fastspeech}. Multi-speaker capabilities are often achieved by conditioning the decoder on speaker embeddings (one-hot embeddings or ones obtained from speaker verification networks~\citep{jia2018transfer}). Training multi-speaker TTS models requires paired text-audio data from multiple speakers. Methods relying on speaker-embeddings can, in theory, perform zero-shot speaker adaptation; however, the rendered speech is known to be of poorer quality, especially for speakers not sufficiently represented in the train set~\citep{tan2021survey}.

\subsection{Raw-audio for TTS}
\label{sec:unsuptts}
Unsupervised speech synthesis~\cite{ni2022unsupervised}  does not require transcribed text-audio pairs for training. 
They typically employ unsupervised ASR~\citep{baevski2021unsupervised,liu2022towards} to transcribe raw speech to generate pseudo labels. However, their performance tends to be bounded by the performance of the unsupervised ASR model, which still has to close a significant gap compared to supervised counterparts~\cite{baevski2021unsupervised}. 
Switching to a multi-speaker setup further widens this quality gap~\cite{liu2022simple}.
% Furthermore, switching to a multi-speaker setup worsens quality relative to fully supervised models~\cite{liu2022simple}.

Some prior works have looked at adapting TTS to novel speakers using untranscribed audio~\cite{yan2021adaspeech,luong2019unified,taigman2017voiceloop}. 
Unlike ours, their methods require a large amount of paired data from multiple speakers during initial training. 
Some of these~\cite{luong2019unified,taigman2017voiceloop} jointly train the TTS pipeline and the modules for speaker adaptation but model training's convergence is trickier. In contrast, \ourmodel~benefits from the disentanglement of linguistic content from speaker information, making adaptation easier with stabler training as we observe in our experiments.

\subsection{Self-supervised learning}
\label{sec:selftts}
% Self-supervised learning (SSL) have helped to make significant advances for applications in computer vision~\cite{he2022masked}, natural language processing~\cite{devlin2018bert,vaswani2017attention} and audio processing~\cite{schneider2019wav2vec}.

% Self-supervised learning (SSL) methods have become increasingly popular in speech processing owing to their ability to leverage copious amounts of unlabeled data. Mask prediction, temporally contrastive learning, next-step prediction, etc., are some common techniques to train SSL models.
% Wav2vec2~\cite{baevski2020wav2vec}, VQ-VAE~\cite{}, AudioLM~\cite{} and Hubert~\cite{hsu2021hubert} are popular SSL models for speech processing and
% ASR~\cite{baevski2020wav2vec}, phoneme segmentation~\cite{kreuk2020self},  and spoken language modeling~\cite{lakhotia2021generative}, speech resynthesis~\cite{polyak2021speech} are tasks that gained from leveraging them. Our work relates to recent efforts~\cite{du2022vqtts,wang2023neural,siuzdak2022wavthruvec} utilizing self-supervised audio embeddings in text-to-speech synthesis. However, SSL embeddings used in these methods~\cite{du2022vqtts,siuzdak2022wavthruvec} are speaker specific, requiring large amount of paired data from multiple speakers for training the TTS model. In contrast, we rely on disentangled Hubert embeddings~\cite{hsu2021hubert,lee2021textless} and combined with the proposes decoupled training procedure allows training a multi-speaker TTS using transcribed data from a single speaker. 

Self-supervised learning (SSL) methods are becoming increasingly popular in speech processing due to their ability to utilize abundant unlabeled data. Techniques like masked prediction, temporally contrastive learning, and next-step prediction are commonly used to train SSL models. 
Popular models like Wav2vec2~\cite{baevski2020wav2vec}, VQ-VAE~\cite{van2017neural}, AudioLM~\cite{borsos2022audiolm} and HuBERT~\cite{hsu2021hubert} have been successfully deployed in tasks like ASR~\cite{baevski2020wav2vec}, phoneme segmentation~\cite{kreuk2020self}, spoken language modeling~\cite{lakhotia2021generative}, and speech resynthesis~\cite{polyak2021speech}.

% Wav2vec2~\cite{baevski2020wav2vec}, VQ-VAE~\cite{van2017neural}, AudioLM~\cite{borsos2022audiolm} and Hubert~\cite{hsu2021hubert} are well-known SSL models in speech processing. They have been beneficial for tasks such as ASR~\cite{baevski2020wav2vec}, phoneme segmentation~\cite{kreuk2020self}, spoken language modeling~\cite{lakhotia2021generative}, and speech resynthesis~\cite{polyak2021speech}.

Our work is related to recent efforts~\cite{du2022vqtts,wang2023neural,siuzdak2022wavthruvec} that utilize self-supervised audio embeddings in text-to-speech synthesis. However, those of~\citet{du2022vqtts} and~\citet{siuzdak2022wavthruvec} require speaker-specific SSL embeddings while we use generic HuBERT embeddings~\cite{hsu2021hubert,lee2021textless} train for multiple speakers.
% and a large amount of paired data from multiple speakers.
% On the contrary, we use generic HuBERT embeddings~\cite{hsu2021hubert,lee2021textless} from multiple speakers trained only on speech data.
% This module is decoupled from the one that handles text using paired data converging well with much fewer hours from single speaker while generalizing well to other speakers.
% and decouple it from training the content module that procedure to train a multi-speaker TTS using transcribed data from a single speaker.

\subsection{Multi-lingual TTS}
\label{sec:multitts}

It is challenging to build an unified TTS model supporting multiple languages and speakers, even more so for cross lingual synthesis, \textit{i.e.}, allowing multiple languages to be spoken in each of the speaker’s voices. 
% Primary challenges to this are, (a) issues curating data (scaling to thousands of languages is significant time and cost expense) and (b) language-dependence of input representations or model's components.
The primary challenge is in acquiring paired data to train language dependent components that often includes its embeddings.
The trick \ourmodel~employs to break this data dependence is to decouple acoustics from content handling, of which only the latter is language dependent and requires paired data while the former is deferred to self-supervised models.

Initial attempts~\cite{liu2019cross,zhang2019learning} address these by conditioning the decoder on language and speaker embeddings, but the results were subpar due to entanglement of text representation with language/speaker information. Recent approaches~\cite{zhang2019learning,cho2022sane,nekvinda2020one} addressed this issue by incorporating an explicit disentanglement loss term, using reverse gradients through a language or speaker classification branch.

% Nekvinda and Dušek~\cite{nekvinda2020one} propose MetaTTS, which uses a contextual parameter generation using language-specific convolutional text encoders. Cho~\etal~\cite{cho2022sane} build upon MetaTTS by employing an additional speaker regularization loss. They also investigate the effect of using different formats for the textual input (phoneme, character, etc.). Some efforts have explored aspects of knowledge sharing~\cite{prakash2019building} and knowledge distillation~\cite{xu2020lrspeech} for multi-lingual TTS. More recently, Wu~\etal~\cite{wu2022multi-lingual} uses a data augmentation technique based on a pre-trained voice conversion model to generate insufficient paired data for a pre-selected set of languages.

%Nekvinda and Dušek~ 
% Cho~\etal~
% Wu et al.~\etal~
\citet{nekvinda2020one} propose MetaTTS, that uses a contextual parameter generation through language-specific convolutional text encoders. \citet{cho2022sane} extend MetaTTS with a speaker regularization loss and investigate different input formats for text. Knowledge sharing~\cite{prakash2019building} and distillation~\cite{xu2020lrspeech} have been explored for multi-lingual TTS. Recently,~\citet{wu2022multilingual} employ a data augmentation technique based on a cross-lingual voice conver-
sion model trained with speaker-invariant features extracted
from a speech representation.

% Certain limitations still persist. For instance, the majority of them~\cite{nekvinda2020one,chen2019cross,zhang2019learning,zhang2020unsupervised} rely on Tacotron~\cite{wang2017tacotron} as their backbone, which is known to suffer from word alignment and word skipping/repetition errors. As pointed out by Nekvinda and Dušek~\cite{nekvinda2020one} most prior multi-lingual TTS models limit to 2-3 languages simultaneously or require vast amounts of data to be trained. Furthermore, they have not fully capitalized on the self-supervised language embeddings and our efforts fill this gap.

Certain limitations still persist in existing approaches~\cite{nekvinda2020one,chen2019cross,zhang2019learning,zhang2020unsupervised}. For example, many of them rely on Tacotron~\cite{wang2017tacotron} as their backbone, which is prone to word alignment and repetition errors. Prior multi-lingual TTS models typically support only 2-3 languages simultaneously or require extensive training data as noted by~\citet{nekvinda2020one}. Additionally, they have not yet capitalized on self-supervised embeddings and our efforts aim to address this gap.

\begin{figure*}[t]
  \centering
  \includegraphics[width=0.95\linewidth]{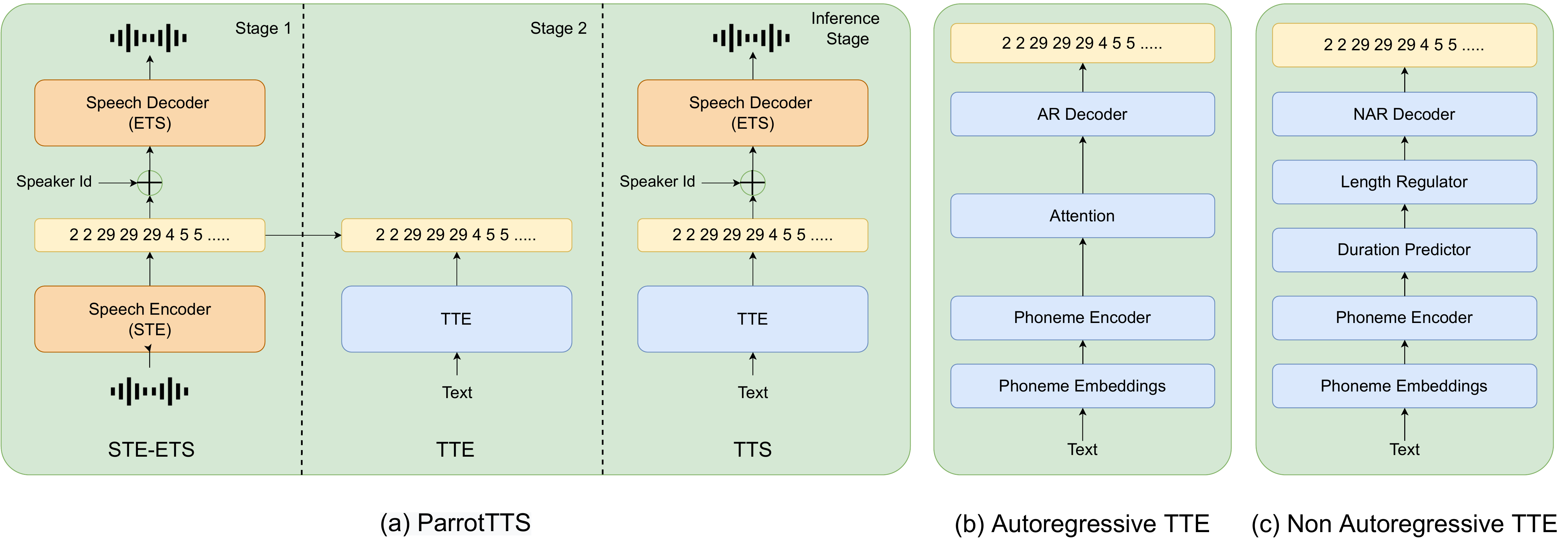}
  \caption{(a) \ourmodel~performs a two stage training. In stage1, ETS is trained to synthesize speech from discrete units obtained though an independently trained STE module. In Stage2, TTE learns to map text sequence to corresponding speech units obtained from STE. (b) and (c) illustrate the explored TTE architectures.  }
  \label{fig:tth}
\end{figure*}

% \begin{figure}[t]
%   \centering
%   \includegraphics[width=\linewidth]{images/stages.png}
%   \caption{Schematic diagram of the proposed model.}
%   \label{fig:tth}
% \end{figure}

% \begin{figure}[t]
%   \centering
%   \includegraphics[width=\linewidth]{images/ar_nar.png}
%   \caption{Schematic diagram of the proposed model.}
%   \label{fig:tth}
% \end{figure}

\section{\ourmodel~architecture}
\label{sec:model}
% To leverage large amounts of speech data 
% In this section we describe the model overview and training pipeline. 
\ourmodel~has three modules; two encoders that map speech or text inputs to common embedding space (referred to as \spencoder~and \txencoder~respectively) and a decoder (\spdecoder) that renders speech signal from these embeddings.
Our speech encoder-decoder choices are borrowed from~\cite{polyak2021speech}.
%The speech encoder \spencoder~is HuBERT~\cite{hsu2021hubert} or mHuBERT~\cite{lee2021textless}.
Our speech decoder \spdecoder~is a modified version of HiFiGAN~\cite{kong2020hifi}.
Text encoder \txencoder~is an encoder-decoder architecture and we experiment with both autoregressive (\ar) and non-autoregressive (\nar) choices for the same. 
%We give architectural details of these modules below.

\subsection{Speech encoder \spencoder} \label{sec:spencoder}
The self-supervised HuBERT model we use for our \spencoder~is pre-trained on large raw audio data from English, on BERT-like masked prediction task~\cite{devlin2018bert} to learn ``combined acoustic and language model over the continuous inputs'' of speech.
It borrows the base architecture from Wav2vec 2.0~\cite{baevski2020wav2vec} with convolutions on raw inputs followed by a few transformer layers, however, replaces its contrastive loss with a BERT-like classification.
The ``noisy'' classes for this classification are derived by clustering MFCC features of short speech signals.
Encoder input is audio signal \(X = (x_1,....x_T)\) sampled at a rate of $16$\khz.
%~through time instance $T$. 
Let $E_r$ denote the raw-audio encoder, and its output be,
\[
\hmb{r} = (h_1,....,{h_{\widehat{T}}}) \coloneqq E_r (X),
\] 
where \(\widehat{T} = T/320\) indicates downsampling and each \(h_i \in \{1,\dots,K\}\) with $K$ being the number of clusters in HuBERT's clustering step, set to $100$ in our experiments. For multi-lingual experiments, instead of using HuBERT, we utilize mHuBERT~\cite{lee2021textless}, which is trained on a multi-lingual corpus. We use $K$=$1000$ for mHuBERT embeddings.  

% For multi-lingual experiments, we instead use mHuBERT~\cite{lee2021textless}, trained on multi-lingual corpus. 

\subsection{Speech decoder \spdecoder}\label{sec:spdecoder}
We adapt the HiFiGAN-v2 vocoder for our \spdecoder~to decode from~$\hmb{} = (\hmb{r}, \hmb{s})$ to speech, where $\hmb{s}$ is the one-hot speaker embedding.
It has a generator $G$ and a discriminator $D$.
$G$ runs~$\hmb{}$ through transposed convolutions for upsampling to recover the original sampling rate followed by residual block with dilations to increase the receptive field to synthesize the signal, $\widehat{X} \coloneqq G(\hmb{})$.

%~\cite{su2021hifi}

The discriminator distinguishes synthesized~$\widehat{X}$ from the original signal~$X$ and is evaluated by two sets of discriminator networks. 
Multi-period discriminators operate on equally spaced samples, and multi-scale discriminators operate at different scales of the input signal.
Overall, the model attempts to minimize~$D(X,\widehat{X})$ over all its parameters to train \spdecoder.

\subsection{Text encoder \txencoder}
The third module we train, \txencoder~is a text encoder that maps phoneme/character sequence~$P = (p_1, \dots, p_N)$ to embedding sequence~$\hmb{p} = (h_1, \dots, h_{\widehat{N}})$.
We train a sequence-to-sequence architecture to achieve this~$\hmb{p} \coloneqq E_p(P)$.
$E_p$ initially encodes~$P$ into a sequence of fixed dimensional vectors (phoneme embeddings), conditioned upon which its sequence generator produces variable dimensional~$\hmb{p}$.
Embedding~$\hmb{p}$ is intended to mimic~$\hmb{r} \coloneqq E_r(X)$ extracted from the audio $X$ corresponding to the text $P$.
Hence, the requirement of transcribed data $(X,P)$ to derive the target~$\hmb{r}$ for training \txencoder~by optimizing over the parameters of~$E_p$.
%to drive down its gap to~$E_p(P)$.
% The length of $\hmb{p}$, $\widehat{N}$ varies with the length of the input phoneme sequence~$N$.

One could model~$E_p$ to generate~$\hmb{p}$ autoregressively one step at a time, which we refer to as \ar-\txencoder~model (Figure~\ref{fig:tth}(b)).
Input phoneme sequence is encoded through a feed-forward transformer block that stacks self-attention layers~\cite{vaswani2017attention} and 1D convolutions similar to FastSpeech2~\cite{ren2019fastspeech}.
Decoding for~$\hmb{p}$ uses a transformer module with self-attention and cross-attention.
Future-masked self-attention attends to ground truth at train and to previous decoder predictions at inference.
Cross-attention attends to phoneme encoding in both cases.

Alternatively, for a non-autoregressive choice of~$E_p$, the model \nar-\txencoder~determines the output length $\widehat{N}$ followed by a step to simultaneously predict all $\widehat{N}$ entries of~$\hmb{p}$.
Figure~\ref{fig:tth}(c) depicts \nar-\txencoder~where the input phoneme sequence encoding is similar to that of~\ar-\txencoder.
The duration predictor and length regulator modules are responsible for determining $\widehat{N}$ followed by the decoding step to generate~$\hmb{p}$. In multi-lingual scenario, we investigate both character and phoneme sequences for representing the input text. For character representation, we extract the tokens using a dictionary created by iterating over the entire text corpus. In contrast, for phoneme representation, we utilize an off-the-shelf phonemizer (version: $3.2.1$) \cite{bernard2021phonemizer} to extract phonemes belonging to the IPA vocabulary, which are common across languages.

\section{Experiments}
\label{sec:experiments}
We perform experiments in monolingual and multi-lingual scenarios.
Details of various \ourmodel~models trained and of those each of them is compared to is covered below.

%We compare against baselines and other comparable models in the literature.

% We train multiple models of the~\ourmodel~under different settings and benchmark them against comparable models in the literature.

% Specifically, we train single-speaker and multi-speaker models to evaluate naturalness, intelligibility, and speaker adaptability. 
% Naturalness is measured by mean-opinion scores (MOS) from human judgments.
% Intelligibility is measured by word-error rates from 
% %using state-of-the-art 
% an ASR model 
% %Whisper~\cite{radford2022robust}  
% on the rendered speech output. Speaker adaptability is measured using Equal-Error-Rate from a pre-trained speaker verification system. 
% We perform these experiments with both autoregressive and non-autoregressive choices of \txencoder.

%about 13k high-quality English transcribed audio clips totaling about
\subsection{\ourmodel~training}
\textbf{Datasets (monolingual)} For single language experiments, we use two public datasets. LJSpeech~\cite{ljspeech17} provides $24$ hours high quality transcribed data from a single speaker. Data are split into two, with $512$ samples set aside for validation and the remaining available for model training. 
VCTK~\cite{Veaux2017CSTRVC} with about $44$ hours of transcribed speech from $108$ different speakers is used for the multi-speaker setup.
It has a minimum, average, and maximum of $7$, $22.8$, and $31$ minutes per speaker speech length, respectively.
%All audio samples are resampled to $16$\khz~ before use.

\textbf{Datasets (multi-lingual)} We collate our multi-lingual dataset using publicly available corpora containing samples from multiple speakers in six languages: (1) $80.76$ hours of Hindi and Marathi from~\cite{SYSPIN} from $2$ speakers, respectively; (2) $71.69$ hours of German \cite{MAILabs} from $3$ speakers; (3) $83.01$ hours of Spanish \cite{MAILabs} from $3$ speakers; (4) $10.70$ hours of French \cite{honnet2017siwis} from $1$ speaker; (5) $23.92$ hours of English \cite{ljspeech17} from $1$ speaker. Overall the dataset comprises of $354.12$ hours of paired TTS data from $12$ speakers across all six languages. We resample all speech samples to $16$~$\khz$. 

\textbf{\spencoder~training.}
We use a $12$ layer transformer model for HuBERT training. It is trained using $960$ hour-long LibriSpeech corpus~\cite{panayotov2015librispeech}. The multi-lingual variant mHuBERT is trained using $13.5$k hours of English, Spanish and French data from VoxPopuli unlabelled speech corpus \cite{lee2021textless,wang2021voxpopuli}. In both cases, the model splits each $T$ seconds long audio into units of $T/320$ seconds and maps each of the obtained units to a $768$ dimensional vector.

% For single language experiments we use $960$ hour-long LibriSpeech corpus~\cite{panayotov2015librispeech} as our \spencoder~module to extract~$\hmb{r}$ embeddings.

% The model splits each $T$ seconds long audio into units of $T/320$ seconds and maps each of the obtained units to a $768$ dimensional vector.
% The vectors are drawn from the network's activation units on the sixth layer similar to that of \citet{lakhotia2021generative}.
% Continuous vectors are then discretized to~$\hmb{r}$ embeddings using a codebook made from applying $k$-means (with $k$ set to $100$) to $100$ hour subset of the data called LibriSpeech-clean~\cite{panayotov2015librispeech}.

% We use a pretrained $12$-layer transformer-based mHuBERT \cite{lee2021textless} architecture trained for $3$ iterations on $13.5$k hours of English, Spanish and French data from VoxPopuli unlabelled speech corpus \cite{lee2021textless,wang2021voxpopuli}. We extract a $768$ dimensional continuous vector from the eleventh activation layer similar to \cite{lee2021textless}. These vectors are further discretized to~$\hmb{s}$ units using a k-means clustering algorithm with $k=1000$.

\textbf{\txencoder~training (monolingual).} We use LJSpeech to train two different \txencoder~encoder modules; \txencoder${}_{\textsc{LJS}}$ that uses all the data from our LJSpeech train set and a second, \txencoder${}_{\frac{1}{2}\textsc{LJS}}$ with only half the data. This is used to understand the effect of training data size on TTS performance. All variants of \txencoder~we experiment with are trained only on samples from the single speaker in LJSpeech data.

Text converted to phoneme sequence as described by~\citet{sun19c_interspeech} are inputs with~$\hmb{r}$ targets extracted from \spencoder~for training. Additionally, \nar-\txencoder~requires phonetic alignment to train the duration predictor. We use Montreal forced-aligner~\cite{mcauliffe2017montreal} to generate them for its training. We use cross-entropy loss with the $100$ clusters derived from discretization codebook of HuBERT units as classes.

%Unlike standard TTS systems that predict Mel spectrograms, \txencoder~generates discrete units. Hence, we replace the mean-square error loss used in Mels with 

\textbf{\txencoder~training (multi-lingual).}  Focusing on low-resource setting, we use only $5$ hours of paired data for a single speaker in each language to train the \txencoder~that totals to merely 30 hours of paired data across all languages. We report the evaluation metrics for \textit{seen speakers} where the model has seen the speaker paired data and \textit{unseen speakers} whose paired data is not used to train the \txencoder.
To evaluate the performance on various text representations, we train two variants of the \txencoder~, the character \txencoder~(CTE) and the phoneme \txencoder~(PTE). CTE uses character tokens across the languages to learn sound units while PTE uses phoneme tokens. Additionally, we employ Deep Forced Aligner \cite{DFA} to align ground-truth speech and input text representations to train the duration predictor. Cross-entropy loss with $1000$ clusters of mHuBERT are used as classes to predict~$\hmb{p}$.  

\textbf{\spdecoder~training.}
We train a single-speaker \spdecoder, SS-\spdecoder~using only speech clips from LJSpeech since its training does not require transcriptions.
Similarly, our multi-speaker \spdecoder, MS-\spdecoder~decoder model uses only raw audio of all speakers from VCTK data~\cite{Veaux2017CSTRVC}.
So only embeddings~$\hmb{r}$ extracted from VCTK audio clips are used along with one-hot speaker vector~$\hmb{s}$. We emphasize that VCTK data were used only in training the multi-speaker-\spdecoder~module, and the \txencoder~has not seen any from this set. For multi-lingual scenario, we train a multi-speaker \spdecoder~using speech-only data with $12$ speakers from all six languages.

\subsection{Comparison to prior art}

\textbf{Single Speaker TTS:} 
We train Tacotron2~\cite{wang2017tacotron} and FastSpeech2~\cite{ren2020fastspeech} using the ground truth transcripts of LJspeech and referred to as SS-Tacotron2 and SS-FastSpeech2. We additionally trained an unsupervised version of FastSpeech2 by replacing the ground truth transcripts with transcriptions obtained from the ASR model. FastSpeech2-SupASR uses supervised ASR model~\cite{radford2022robust} to generate the transcripts while Tacotron2-UnsupASR~\cite{ni2022unsupervised} alternatively uses unsupervised ASR Wav2vec-U 2.0 \cite{liu2022towards}. We further adapt WavThruVec~\cite{siuzdak2022wavthruvec} to our setup and train a model (SS-WavThruVec) using intermediate embeddings extracted from Wav2Vec 2.0~\cite{baevski2020wav2vec}. Additionally, we apply a similar approach to the embeddings obtained from VQ-VAE \cite{van2017neural} and term it as SS-VQ-VAES.
% with a Tacotron backbone with Transformers described by~\citet{ni2022unsupervised}.
We compare against three variants of \ourmodel; 
\begin{enumerate}
    \item  \ar-\txencoder${}_{\text{LJS}}$+SS-\spdecoder~that is autoregressive \txencoder~trained on full LJSpeech with single speaker \spdecoder,
    
    \item \nar-\txencoder${}_{\text{LJS}}$+SS-\spdecoder~that pairs \txencoder~with non-autoregressive decoding trained on full LJSpeech with single speaker \spdecoder, and
    
    \item \nar-\txencoder${}_{\frac{1}{2}\text{LJS}}$+SS-\spdecoder~that uses \txencoder~with non-autoregressive decoding trained on half LJSpeech with single speaker \spdecoder.
\end{enumerate}
% Fastspeech2\cite{ren2020fastspeech}. We train two variants of Fastspeech2, one using ground truth transcripts and another where the transcripts are . We also compare against an Unsupervised TTS model~\cite{ni2022unsupervised} trained which is trained on transcripts generated from unsupervised ASR model Wav2vec-U 2.0 \cite{liu2022towards}.

% We compare the above models against variants of \ourmodel categorized by either the choice of architecture or the amount of transcribed data used in \txencoder~training. 
% We train \txencoder~module autoregressively (AT-TTE) and non-autoregressively (NAT-TTE) using ground truth transcripts of LJSpeech dataset. For NAT-TTE, we also experiment with training the TTE module using half of the transcribed data.

\noindent
\textbf{Multi-speaker TTS:} 
We compare against a fully supervised Fastspeech2 baseline trained on VCTK using paired data from all speakers and that we refer to as MS-FastSpeech2. 
%In contrast, we do not use even a single line of transcripts from the VCTK data for any of our \ourmodel variants we train. 
For \ourmodel~we borrow the \txencoder~module trained on LJSpeech and use the raw audio of VCTK to train the multi-speaker \spdecoder~module. 
We refer to this multi-speaker variant of our \ourmodel~model as \nar-\txencoder${}_{\text{LJS}}$+MS-\spdecoder~that uses non-autoregressive decoding.

%for \txencoder~similar to the FastSpeech2 baseline trained on LJSpeech alone and multi-speaker \spdecoder~trained on VCTK alone.

For a fair comparison, we also curate a multi-speaker TTS baseline using a combination of single-speaker TTS and a voice cloning model. 
We use FastSpeech2 trained on LJspeech with state-of-the-art voice cloning model~\cite{polyak2021speech} in our experiments and refer to this model as VC-FastSpeech2. We also compare against multi-speaker TTS trained by obtaining pseudo labels from a supervised ASR called MS-FastSpeech2-SupASR.
%and an unsupervised ASR. 
Additionally, we also report numbers from GT-Mel+Vocoder that converts ground truth Mels from actual audio clip back to speech using a vocoder~\cite{kong2020hifi} for a perspective of best achievable with ideal Mel frames.

\noindent
\textbf{Multi-lingual TTS:} We compare against, (a) FastSpeech2-MLS which is a fully-supervised FastSpeech2 model and (b) state-of-the-art meta learning-based multi-lingual TTS model \baselineml~\cite{nekvinda2020one}. Both these models are trained on the entirety of train data ($354$ hours of transcribed speech).
In contrast, the \txencoder~training in \ourmodel~model (our sole module that needs paired data) uses only $1/12^{th}$ of this \textit{i.e}, a total of $30$ hours of paired text-speech (5 hours per language). The remaining data is used for evaluation purposes, serving as the test set. We refer to this model as NAR-TTE$_{\frac{1}{12}\text{MLS}}$+ML-ETS. We also compare character (CTE) and phoneme (PTE) tokenization for encoding text in this setting.

\subsection{Evaluation metrics}
We evaluate the intelligibility of various models using Word Error Rate (WER) with the pre-trained Whisper \textit{small} model~\cite{radford2022robust}. We validate the speaker adaptability using Equal Error Rate (EER) from a pre-trained speaker verification network proposed in~\cite{desplanques2020ecapa} and trained on VoxCeleb2~\cite{chung2018voxceleb2}. The WER and EER metrics are computed on entire validation set. We perform subjective evaluations using Mean Opinion Score (MOS) with five native speakers per language, rating samples synthesized by different models, where five sentences from the test set are randomly selected for evaluation.

\section{Results}
\label{sec:results}
% Quantitative and qualitative results evaluating the proposed ParrotTTS system are shown in Tables~\ref{tab:tts-mos} and~\ref{tab:tts-vctk} for single-speaker and multi-speaker models, respectively. 

\subsection{Single-speaker TTS}

\textit{Naturalness and intelligibility.} As shown in Table~\ref{tab:tts-mos}, \ourmodel~is competitive to state-of-the-art in the single-speaker setting.
In the autoregressive case, our \ar-\txencoder${}_{\textsc{LJS}}$+SS-\spdecoder~ has a statistically insignificant drop (of about $0.05$ units) on the MOS scale relative to the Tacotron2 baseline.
The non-autoregressive case has a similar observation (with a $0.01$ drop) on MOS in our \nar-\txencoder${}_{\textsc{LJS}}$+SS-\spdecoder~model relative to FastSpeech2.
This empirically establishes that the naturalness of the speech rendered by \ourmodel~is on par with the currently established methods.
The WER scores show a similar trend with a statistically insignificant drop (of under $0.2$pp\footnote{Percentage points abbreviated as pp.}) among the autoregressive and non-autoregressive model classes. The performance of SS-WavThruVec and SS-VQ-VAES is lower in both naturalness and intelligibility, indicating that the utilization of Wav2Vec 2.0 and VQ-VAE embeddings results in a decrease in performance.

\textit{Supervision and data efficiency.} In the study to understand how the degree of supervision affects TTS speech quality, we see a clear drop by $0.28$ MOS units in moving from the FastSpeech2-SupASR model that employs supervised ASR for transcriptions to Tacotron2-UnsupASR model using unsupervised ASR. Despite some modeling variations, this is generally indicative of the importance of clean transcriptions on TTS output quality, given that all other models are within $0.05$ MOS units of each other.

The data requirement for TTS supervision needs to be understood in light of this impact on output quality, and we show how \ourmodel~helps cut down on this dependence.
\txencoder~is the only module that needs transcriptions to train, and we show that by reducing the size of the train set by half in \nar-\txencoder${}_{\frac{1}{2}\textsc{LJS}}$+SS-\spdecoder~the MOS is still comparable to that of the model trained on all data \nar-\txencoder${}_{\textsc{LJS}}$+SS-\spdecoder~(with only about $0.04$ units MOS drop).
Finally, the MOS numbers of FastSpeech2-SupASR, need to be read with some caution since the supervised ASR model used, Whisper, is itself trained with plenty of transcriptions (spanning over $600$k hours) from the web, including human and machine transcribed data achieving very low WERs on various public and test sets. So, the machine transcriptions used in FastSpeech2-SupASR are indeed close to ground truth. 
%with a WER of~$2.3$\%..

\begin{table}[t]
\scriptsize
\centering
\begin{tabular}{llccc}%llllllll
% \begin{subtable}[t]{.55\textwidth}
\hline
&\textbf{Model} &\textbf{MOS} $\uparrow$ &\textbf{WER} $\downarrow$  \\
% \hline
% & &&Happy&Sad &Angry &Fear&Average& 
\hline
\multirow{4}{*}{Traditional TTS}&
SS-FastSpeech2 & 3.87 &4.52  \\
&SS-Tacotron2 & 3.90 &4.59 \\
&FastSpeech2-SupASR & 3.78&4.72\\
&Tacotron2-UnsupASR & 3.50&11.3\\
\hline
\multirow{1}{*}{WavThruVec}
&SS-WavThruVec& 3.57 & 6.27\\
\hline
\multirow{1}{*}{VQ-VAE}
&SS-VQ-VAES& 3.12 & 21.78\\
\hline
\multirow{3}{*}{ParrotTTS}
&\ar-\txencoder${}_{\text{LJS}}$+SS-\spdecoder~&3.85&4.80\\
&\nar-\txencoder${}_{\text{LJS}}$+SS-\spdecoder~&3.86&4.58\\
&\nar-\txencoder${}_{\frac{1}{2}\text{LJS}}$+SS-\spdecoder~&3.81&6.14\\
% NAR-TTH (Blizzard)&3.65&7.52\\
% NAR-TTH (Iemocap)&3.6&x\\
% \vspace{-5mm}

\hline
\end{tabular}
\caption{\label{tts-mos}
Subjective and objective comparison of TTS models in the single speaker setting.
}
\label{tab:tts-mos}
\end{table}

\begin{table}[t]
\scriptsize
% \begin{subtable}
\centering
\begin{tabular}{lcccc}%llllllll
\hline
\textbf{Model} & \textbf{VCTK} & \textbf{MOS} $\uparrow$ &\textbf{WER} $\downarrow$
&\textbf{EER} $\downarrow$ \\
% \hline
% & &&Happy&Sad &Angry &Fear&Average& 
\hline
GT-Mel+Vocoder & Yes & 4.12 &2.25 & 2.12\\
MS-FastSpeech2 & Yes & 3.62 &5.32 & 3.21 \\
MS-FastSpeech2-SupASR & No & 3.58 & 6.65 & 3.85 \\
%Unsupervised TTS (Unsupervised ASR) & No & 3.42 &13.56 & 4.23 \\

VC-FastSpeech2 & No & 3.41 &7.44 & 8.18 \\
WavThruVec-MS & No & 3.17 & 6.79 & 5.08 \\
\hline

\nar-\txencoder${}_{\text{LJS}}$+MS-\spdecoder~ & No & 3.78 &6.53 & 4.38\\
% AR-TTH (LJSpeech) & 3.54&x\\

\hline
\end{tabular}
\caption{\label{tts-mos}
Comparison of the multi-speaker TTS models on the VCTK dataset. Column $2$ indicates if the corresponding method uses VCTK transcripts while training. 
}
\label{tab:tts-vctk}
% \end{subtable}
\end{table}

% \begin{figure}[t]
%   \centering
%   \includegraphics[width=\linewidth]{images/steps.png}
%   \caption{Visualization of attention between output units and phonemes. (a) Evolution of attention matrix with training steps. (b) Attention loss plotted against training steps.} 
  
%   \label{fig:attentions}
% \end{figure}

\subsection{Multi-speaker TTS}
\textit{Naturalness and intelligibility.} 
Table~\ref{tab:tts-vctk} summarizes results from our multi-speaker experiments. \nar-\txencoder${}_{\textsc{LJS}}$+MS-\spdecoder~clearly outperforms all other models ranking only below GT-Mel+Vocoder that re-synthesizes from ground truth Mels.
Interestingly, \ourmodel~fares even better than MS-FastSpeech2, which is, in turn, better than other models that ignore transcripts at the train, namely, MS-FastSpeech2-SupASR and VC-FastSpeech2.
On the WER metric for intelligibility, \ourmodel~is about $1$pp behind supervised MS-FastSpeech2 but fares better than the other two models that discard VCTK transcripts for training. WavThruVec-MS model leveraging Wav2Vec 2.0 embeddings has a noticeable quality drop in the multi-speaker setting with lowest MOS. 

\textit{Speaker adaptability.}
VC-FastSpeech2 is the closest in terms of experimental setup since it is limited to transcriptions from LJSpeech for training similar to ours, with VCTK used only for adaptation. In this case, EER of \nar-\txencoder${}_{\textsc{LJS}}$+MS-\spdecoder~is about twice as good as that of VC-FastSpeech2. However, improvements are visible when VCTK transcripts are part of training data but remain within $1$pp relative to \ourmodel~while GT-Mel+Vocoder continues to dominate the scoreboard leaving room for further improvement.
%\textcolor{red}{But we use VCTK audio for speakers, they do it as well?}

\subsection{Multi-lingual TTS}
The results from our multi-lingual experiments are in Tables~\ref{tab:mos-naturalness-seen},~\ref{tab:mos-naturalness-unseen},~\ref{tab:mos-naturalness-speakersimilarity}, 
 and \ref{tab:mos-naturalness-crosslingual}.
It is notable that speech rendered by \ourmodel~has superior naturalness compared to baselines that are trained with twelve times more paired samples stressing its viability for low-resource languages.
Further, the naturalness also changes with the text tokenization method. 
Choosing character tokens for Indic languages outperformed phoneme tokens while it was the opposite for the European languages. 
\ourmodel~with the best performing tokeniser in each language was superior to FastSpeech2-MLS and \baselineml~for both \textit{seen speakers} (Table~\ref{tab:mos-naturalness-seen}) as well as \textit{unseen speakers} (Table~\ref{tab:mos-naturalness-unseen}).
It is interesting to note that scores for \ourmodel~were better than groundtruth and this is possibly due to noise in original sample that was suppressed by HuBERT embeddings that are known to discard ambient information.% This is possibly due to poor quality of training samples where even the groundtruth scored a low MOS.

\textit{Speaker similarity.} %We conducted a comparative analysis to evaluate the ability of \ourmodel~to capture speaker identity. Speaker similarity is reported on a scale of $1$ to $5$, where $1$ corresponds to ``Not at all similar,'' and $5$ corresponds to ``extremely similar''. 
Results in Table~\ref{tab:mos-naturalness-speakersimilarity} consistently demonstrate the superiority of \ourmodel~over FastSpeech2-MLS and \baselineml, indicating its effectiveness in separating speaker and content information. This is attributed to the decoder being conditioned solely on speaker ID while sharing the acoustic space across all languages.

\begin{table}%[t]
%\scriptsize
  \centering
   \resizebox{0.5\textwidth}{!}{
  \begin{tabular}{ccccccc}
    \hline %\toprule
    \multicolumn{1}{c}{\textbf{}} & 
    \multicolumn{1}{c}{\textbf{GT}} &
    \multicolumn{1}{c}{\textbf{CTE (Ours)}} &
    \multicolumn{1}{c}{\textbf{PTE (Ours)}} &
    \multicolumn{1}{c}{\textbf{FS2-MLS}} &
    \multicolumn{1}{c}{\textbf{\baselineml~}}\\
    \hline %\midrule
    Hindi & $3.78$ $\pm$ $0.14$ & $\textbf{3.33}$ $\pm$ $\textbf{0.19}$ & $3.22$ $\pm$ $0.15$ & $3.33$ $\pm$ $0.12$ & $2.12$ $\pm$ $0.12$\\
    Marathi & $4.81$ $\pm$ $0.07$ & $\textbf{3.78}$ $\pm$ $\textbf{0.12}$ & $3.04$ $\pm$ $0.19$ & $3.59$ $\pm$ $0.15$ & $2.13$ $\pm$ $0.15$\\
    German & $3.54$ $\pm$ $0.20$ & $3.33$ $\pm$ $0.19$ & $\textbf{3.58}$ $\pm$ $\textbf{0.12}$ & $3.21$ $\pm$ $0.16$ & $1.8$ $\pm$ $0.15$\\
    French & $3.83$ $\pm$ $0.19$ & $2.23$ $\pm$ $0.14$ & $\textbf{4.17}$ $\pm$ $\textbf{0.19}$ & $3.50$ $\pm$ $0.16$ & $1.7$ $\pm$ $0.16$\\
    English & $4.20$ $\pm$ $0.12$ & $3.11$ $\pm$ $0.11$ & $\textbf{3.50}$ $\pm$ $\textbf{0.10}$ & $2.50$ $\pm$ $0.18$ & $1.6$ $\pm$ $0.17$\\
    Spanish & $3.67$ $\pm$ $0.12$ & $3.5$ $\pm$ $0.21$ & $\textbf{3.67}$ $\pm$ $\textbf{0.20}$ & $2.50$ $\pm$ $0.21$ & $2.1$ $\pm$ $0.15$\\
    \hline %\bottomrule
  \end{tabular}
  }
  \caption{Comparison of naturalness MOS on seen speakers with FastSpeech2-MLS (FS2-MLS) and \baselineml~model}
 \label{tab:mos-naturalness-seen}
\end{table}

\begin{table}[t]
%\scriptsize
  \centering
   \resizebox{0.5\textwidth}{!}{
  \begin{tabular}{cccccccc}
    \hline %\toprule
    \multicolumn{1}{c}{\textbf{}} & 
    \multicolumn{1}{c}{\textbf{GT}} &
    \multicolumn{1}{c}{\textbf{CTE (Ours)}} &
    \multicolumn{1}{c}{\textbf{PTE (Ours)}} &
    \multicolumn{1}{c}{\textbf{FS2-MLS}} &
    \multicolumn{1}{c}{\textbf{\baselineml~}}\\
    \hline %\midrule
    Hindi & $4.22$ $\pm$ $0.18$ & $\textbf{3.28}$ $\pm$ $\textbf{0.19}$ & $3.05$ $\pm$ $0.20$ & $3.22$ $\pm$ $0.21$ & $2.02$ $\pm$ $0.18$ \\
    Marathi & $4.48$ $\pm$ $0.13$ & $\textbf{3.63}$ $\pm$ $\textbf{0.18}$ & $3.11$ $\pm$ $0.18$ & $3.15$ $\pm$ $0.19$ & $1.91$ $\pm$ $0.19$ \\
    German & $3.17$ $\pm$ $0.22$ & $2.72$ $\pm$ $0.23$ & $\textbf{3.55}$ $\pm$ $\textbf{0.20}$ & $2.05$ $\pm$ $0.22$ & $1.8$ $\pm$ $0.17$ \\
    Spanish & $3.67$ $\pm$ $0.19$ & $3.17$ $\pm$ $0.17$ & $\textbf{3.33}$ $\pm$ $\textbf{0.18}$ & $3.17$ $\pm$ $0.19$ & $1.3$ $\pm$ $0.16$ \\
    \hline %\bottomrule
  \end{tabular}
  }
  \caption{Comparison of naturalness MOS on unseen speakers with FastSpeech2-MLS (FS2-MLS) and \baselineml~model}
  \label{tab:mos-naturalness-unseen}
\end{table}

% \begin{table*}[t]
%   \caption{Comparison of Naturalness MOS with \baselineml~model}
%   \label{tab:mos-naturalness-meta}
%   \centering
%   \begin{tabular}{rrrrrrr}
%     \hline %\toprule
%     \multicolumn{1}{c}{\textbf{Language}} & 
%     \multicolumn{1}{c}{\textbf{Our model}} &
%     \multicolumn{1}{c}{\textbf{\baselineml~}} \\
%     \hline %\midrule
%     French & $\textbf{4.08}$ $\pm$ $\textbf{0.19}$ & $2.58$ $\pm$ $0.18$ \\
%     Spanish & $\textbf{3.75}$ $\pm$ $\textbf{0.16}$ & $3.00$ $\pm$ $0.20$ \\
%     German & $3.33$ $\pm$ $0.19$ & $\textbf{3.50}$ $\pm$ $\textbf{0.18}$ \\
%     \hline %\bottomrule
%   \end{tabular}
% \end{table*}

\begin{table}[t]
\scriptsize
  \centering
  \begin{tabular}{rrrrrrr}
    \hline %\toprule
    \multicolumn{1}{c}{\textbf{Language}} & 
    \multicolumn{1}{c}{\textbf{Our model}} &
    \multicolumn{1}{c}{\textbf{FS2-MLS}} &
    \multicolumn{1}{c}{\textbf{\baselineml~}} \\
    \hline %\midrule
    Hindi & $\textbf{4.29}$ $\pm$ $\textbf{0.18}$ & $3.92$ $\pm$ $0.21$ & $2.23$ $\pm$ $0.19$ \\
    Marathi & $\textbf{4.21}$ $\pm$ $\textbf{0.16}$ & $3.83$ $\pm$ $0.08$ & $2.12$ $\pm$ $0.16$ \\
    German & $\textbf{4.09}$ $\pm$ $\textbf{0.11}$ & $3.25$ $\pm$ $0.14$ & $2.05$ $\pm$ $0.14$ \\
    French & $\textbf{3.87}$ $\pm$ $\textbf{0.20}$ & $3.50$ $\pm$ $0.19$ & $2.24$ $\pm$ $0.17$ \\
    English & $\textbf{3.94}$ $\pm$ $\textbf{0.18}$ & $3.00$ $\pm$ $0.19$ & $2.32$ $\pm$ $0.19$ \\
    Spanish & $\textbf{4.33}$ $\pm$ $\textbf{0.17}$ & $3.50$ $\pm$ $0.19$ & $2.0$ $\pm$ $0.18$ \\
    \hline %\bottomrule
  \end{tabular}
  \caption{Comparison of speaker similarity MOS with FastSpeech2-MLS (FS2-MLS) and \baselineml~model}
    \label{tab:mos-naturalness-speakersimilarity}
\end{table}

\begin{table}[t]
\scriptsize
  \centering
  \begin{tabular}{rrrrrrr}
    \hline %\toprule
    \multicolumn{1}{c}{\textbf{Speaker-Text}} & 
    \multicolumn{1}{c}{\textbf{Our model}} &
    \multicolumn{1}{c}{\textbf{FS2-MLS}} &
    \multicolumn{1}{c}{\textbf{\baselineml~}}\\
    \hline %\midrule
    Hindi-Spanish & $\textbf{3.87}$ $\pm$ $\textbf{0.22}$ & $3.25$ $\pm$ $0.19$ & $1.26$ $\pm$ $0.15$ \\
    Marathi-English & $\textbf{3.63}$ $\pm$ $\textbf{0.21}$ & $3.5$ $\pm$ $0.22$ & $1.23$ $\pm$ $0.19$ \\
    French-Hindi & $\textbf{4.07}$ $\pm$ $\textbf{0.12}$ & $2.71$ $\pm$ $0.21$ & $1.23$ $\pm$ $0.16$ \\
    Spanish-German & $\textbf{4.14}$ $\pm$ $\textbf{0.20}$ & $2.29$ $\pm$ $0.21$ & $1.45$ $\pm$ $0.19$ \\
    English-German & $\textbf{3.57}$ $\pm$ $\textbf{0.15}$ & $2.43$ $\pm$ $0.18$ & $1.56$ $\pm$ $0.16$ \\
    English-Hindi & $\textbf{3.57}$ $\pm$ $\textbf{0.19}$ & $2.57$ $\pm$ $0.18$ & $1.23$ $\pm$ $0.19$ \\
    French-German & $\textbf{3.93}$ $\pm$ $\textbf{0.17}$ & $2.71$ $\pm$ $0.18$ & $1.18$ $\pm$ $0.17$ \\
    Spanish-French & $\textbf{3.71}$ $\pm$ $\textbf{0.18}$ & $2.57$ $\pm$ $0.17$ & $1.4$ $\pm$ $0.16$ \\
    Hindi-Marathi & $\textbf{4.13}$ $\pm$ $\textbf{0.21}$ & $3.25$ $\pm$ $0.19$ & $1.3$ $\pm$ $0.18$ \\
    Marathi-French & $\textbf{2.87}$ $\pm$ $\textbf{0.19}$ & $2.75$ $\pm$ $0.18$ & $1.25$ $\pm$ $0.19$ \\
    \hline %\bottomrule
  \end{tabular}
  \caption{Comparison of naturalness MOS for cross-lingual speech synthesis with FastSpeech2-MLS (FS2-MLS) and \baselineml~model}
    \label{tab:mos-naturalness-crosslingual}
\end{table}

\textit{Cross lingual synthesis.}
We also assess the model's performance in synthesizing samples of a speaker in a language different from native language. 
Table~\ref{tab:mos-naturalness-crosslingual} presents these results comparing naturalness of MOS in a cross-lingual setting. 
The first column lists a pair of languages of which the first is the speaker's native language while the second is language of text that is rendered.
\ourmodel~achieved higher MOS demonstrating strong decoupling of content from speaker characteristics that is controlled in the decoder.
Further, more than $90$\% of the participants were able to discern the nativity of the synthesized speech.

\subsection{Stabler training and faster inference}
We observe that \nar-\txencoder~converges (in $20$k steps) about eight times faster than FastSpeech2 ($160$k steps) during training. Similarly, \ar-\txencoder~model converges 10-times faster than the corresponding Tacotron2 counterpart. The proposed \nar-TTE system also improves inference latency and memory footprint. On NVIDIA RTX $2080$ Ti GPU, we observe \ourmodel~serves $15\%$ faster than FastSpeech2. Furthermore, the \txencoder~module uses $17$M parameters in contrast to $35$M parameters of the Mel synthesizer module in Fastspeech2. More details are provided in the supplementary material.

\section{Conclusion, limitations and future work} 
\label{sec:conclusion}
We investigate a data-efficient \ourmodel~ model that leverages audio pre-training from self-supervised models and ties it to separately trained speech decoding and text encoding modules. We evaluate this architecture in various settings. Quality of rendered speech with as little as five hours of paired data per language is on par with or superior to competitive baselines. This is the key result from our experiments that we believe will help scale TTS training easily to new languages by bringing low-resource ones into the same quality range as the resource-rich ones.

In the future, we plan to fine-tune the Hubert-based embeddings on diverse set of languages (South Asian, Latin, English, etc.) to create a more comprehensive set of sound units. Another direction being to improve upon the data efficiency for speaker adaptability~\cite{wang2023neural}.
Investigations into emotive speech and controllable generation is another aspect. For example, Hubert embeddings are known to skip prosody information~\cite{kharitonov2021text} and hence giving emotive affect to speech would be a challenge in this setup.  Finally, we aim to release an open-source, multi-lingual TTS model to enable the wider application of our findings to resource-scarce and less privileged languages.

\section{Ethical Considerations}
Our research is grounded in ethical considerations. We recognize the potential of text-to-speech synthesis in various domains, such as accessibility, human-computer interaction, telecommunications, and education. However, we acknowledge the risk of misuse, particularly with regards to unethical cloning and the creation of false audio recordings. Our experiments strictly use publicly available datasets and our method does not aim to synthesize someone's voice without their consent. We are mindful of the negative consequences associated with these actions. While the benefits currently outweigh the concerns, we strongly advocate for the research community to actively explore methods for detecting and preventing misuse.

It is important to note that our approach is trained on a limited set of languages and has not been validated on different languages or individuals with speech impediments. Therefore, the dataset and results may not be representative of the entire population. A comprehensive understanding of this issue necessitates further studies in conjunction with linguistic and socio-cultural insights.

% One of the ethical concerns surrounding the development of human-level text-to-speech systems is the potential of the technology to be used for nefarious purposes. For example, someone could use a text-to-speech system to create a fake audio recording of a person saying something they never actually said. The generated voice can also be used to impersonate any individual's voice and use it to spoof an authentication system. However, the benefits of improvements to TTS technology could significantly benefit HCI and the corresponding applications to problems in healthcare and other domains. Although, the current TTS technologies can mimic an individual, they can't capture the exact nuances and unique qualities of a persons voice. While the benefits seem to outweigh the concerns at this point, we believe the research community should proactively continue to identify methods for detection and prevention of misuse. 

\bibliography{custom}
\bibliographystyle{acl_natbib}

\end{document}